\PassOptionsToPackage{table}{xcolor}

\documentclass[10pt,twocolumn,letterpaper]{article}

\usepackage[pagenumbers]{iccv} 
\newcommand{\Tref}[1]{Table~\ref{#1}}

\newcommand{\fref}[1]{Fig.~\ref{#1}}

\newcommand{\sref}[1]{Sec.~\ref{#1}}

\newcommand{\textblock}[1]{\noindent\textbf{#1}}

\def\eg{\emph{e.g}\onedot}

\newcommand{\changed}[1]{{#1}}

\def\ourPaperTitle {ExploreGS: Explorable 3D Scene Reconstruction with Virtual Camera Samplings and Diffusion Priors}

\definecolor{iccvblue}{rgb}{0.21,0.49,0.74}
\usepackage[pagebackref,breaklinks,colorlinks,allcolors=iccvblue]{hyperref}
\usepackage{multirow}
\usepackage{cuted}
\usepackage{xcolor}
\usepackage{algorithm}
\usepackage{algpseudocode}
\usepackage{enumitem}
\usepackage{makecell}
\usepackage{amsmath}
\usepackage{amssymb}
\usepackage{graphicx}
\usepackage{subcaption}
\usepackage{gensymb}
\usepackage{booktabs}
\usepackage{algorithm}
\usepackage{algpseudocode}
\usepackage{textcomp}

\def\paperID{\ourPaperId} 
\def\confName{ICCV}
\def\confYear{2025}

\title{\ourPaperTitle}

\author{
Minsu Kim \quad Subin Jeon \quad In Cho \quad Mijin Yoo \quad Seon Joo Kim\\[1mm]
Yonsei University \\
\href{https://exploregs.github.io/}{\tt\small{ExploreGS.github.io}}
}
\begin{document}

\twocolumn[{
\renewcommand\twocolumn[1][]{#1}%
\maketitle
\vspace{-1 cm}
\begin{center}
    \centering
    \includegraphics[width=\textwidth]{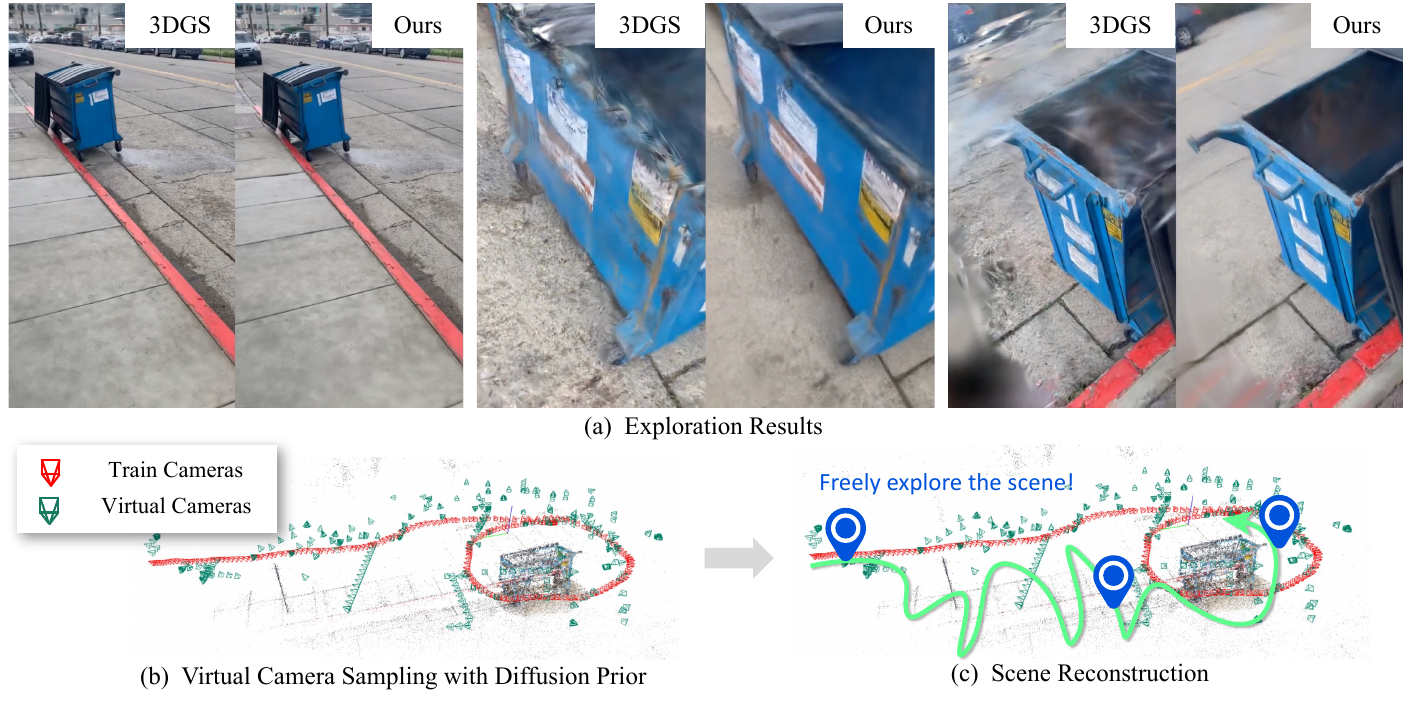}  
    \vspace{-0.7cm} 
    \captionof{figure}{
        (a) Explorable 3D scene reconstruction results. Our method renders photorealistic images from any arbitrary viewpoints. 
        (b) Virtual cameras are sampled in regions with missing information, and pseudo-ground truth is generated using diffusion priors. 
        (c) After optimizing 3D Gaussians with virtual viewpoint samples, our method enables real-time scene exploration.  }
    \label{fig:teaser}
\end{center}
}]

\begin{abstract}
Recent advances in novel view synthesis (NVS) have enabled real-time rendering with 3D Gaussian Splatting (3DGS).
However, existing methods struggle with artifacts and missing regions when rendering from viewpoints that deviate from the training trajectory, limiting seamless scene exploration.
To address this, we propose a 3DGS-based pipeline that generates additional training views to enhance reconstruction.
We introduce an information-gain-driven virtual camera placement strategy to maximize scene coverage, followed by video diffusion priors to refine rendered results.
Fine-tuning 3D Gaussians with these enhanced views significantly improves reconstruction quality.
To evaluate our method, we present Wild-Explore, a benchmark designed for challenging scene exploration.
Experiments demonstrate that our approach outperforms existing 3DGS-based methods, enabling high-quality, artifact-free rendering from arbitrary viewpoints.
\end{abstract}
\section{Introduction}
The field of novel view synthesis (NVS) has seen significant advancements, enabling photorealistic rendering of scenes from previously unseen viewpoints.
Much of this progress is attributed to NeRF \cite{mildenhall2020nerf} and its variants \cite{barron2022mip, muller2022instant}.
This advancement has been further accelerated by recent 3D Gaussian Splatting (3DGS) \cite{kerbl20233d}, which enables high-quality rendering in real-time.
The real-time capability of 3DGS broadens the applications of NVS in domains such as AR, VR, and gaming, leading users to expect the experience of freely exploring reconstructed scenes.

Unfortunately, such an experience is yet to be fully realized, as existing methods suffer from severe degradations in rendering quality when viewpoints deviate significantly from the input observations.
These degradations include holes, oversized Gaussians, spiky surface artifacts, and spurious floaters, often arising from occlusions or limited viewpoint coverage (see \fref{fig:teaser}~(a)).
This limitation stems from missing information, since optimization-based approaches cannot synthesize contents beyond the observed data.

We are interested in 3D scene reconstruction that can render more vivid and photorealistic images from arbitrary viewpoints, so that users can \textit{freely explore the reconstructed scenes.}
We term this task as ``explorable scene reconstruction'', which aims to reconstruct complete scenes within the target areas.
To avoid generating arbitrary contents in regions far beyond the input observations--which would not reflect actual user experiences--we define the target areas as a bounding box defined by the input views.

One straightforward solution is to take thousands of images to completely cover the target scenes.
However, collecting such dense data is time-consuming and labor-intensive, making it impractical for many real-world applications.
Thus, we aim to reconstruct complete 3D scenes from input images that only partially cover the target scenes, which better reflects practical capturing scenarios.

In this paper, we introduce ExploreGS, a pipeline that enables explorable scene reconstruction using diffusion priors and 3DGS.
ExploreGS leverages video diffusion priors to generate additional training views that complement missing information.
These synthesized images at virtual viewpoints serve as pseudo ground truths to refine the initial, incomplete 3D Gaussians, thereby enabling complete, photorealistic rendering at arbitrary viewpoints.

The key challenges of explorable scene reconstruction lie in determining the optimal placement of virtual viewpoints.
Unlike previous works that assume evaluation trajectories are known~\cite{liu2024viewextrapolator, shih2024extranerf}, virtual viewpoints in our pipeline need to be sampled to best complement missing information in poorly covered areas.
We propose a novel view sampling strategy that strategically positions virtual cameras based on information gain.
This approach identifies viewpoints that offer the most novel scene details from the initial 3DGS reconstruction as shown in \fref{fig:teaser} (b).

To provide pseudo observations, we generate realistic images along sampled viewpoints.
We introduce video diffusion model which converts renderings with degradations into clean images.
Although it is motivated by 3DGS-Enhancer \cite{liu20243dgsenhancer}, we modify the training objectives to fit scene exploration and prepare larger dataset for training.
We then finetune 3DGS on both virtual viewpoints and training viewpoints with uncertainty guidance.

To validate our pipeline, we introduce Wild-Explore, a dataset designed to evaluate challenging exploration scenarios with diverse and distant viewpoints. 
We also assess our method on the Nerfbusters dataset \cite{warburg2023nerfbustersremovingghostlyartifacts}, curated to better reflect scene exploration tasks. 
Experimental results show that our approach achieves the highest scores among 3DGS-based methods. 
Finally, an interactive exploration demo demonstrates that our method enables broader scene coverage with fewer artifacts compared to existing approaches.

In summary, our contributions can be organized as follows:
\begin{itemize} 
\item We propose a pipeline for explorable 3D scene reconstruction, which incorporates the real-time rendering of 3DGS, video diffusion priors to complete missing regions, and an uncertainty-based finetuning strategy.
\item We propose a novel virtual camera sampling strategy to identify viewpoints that can maximize additional information gain, thereby effectively reducing artifacts and missing regions.
\item We introduce Wild-Explore, a new benchmark dataset designed to evaluate challenging exploration scenarios.
\end{itemize}
\section{Related work}

\subsection{Novel view synthesis}
Novel view synthesis aims to generate images of a scene from unseen viewpoints.
NeRF \cite{mildenhall2020nerf} and its following works \cite{barron2022mip, barron2023zipnerfantialiasedgridbasedneural} have revolutionized novel view synthesis by representing 3D scenes as continuous volumetric functions, significantly improving rendering quality yet still suffering from slow rendering speed.
Recently, 3D Gaussian Splatting \cite{kerbl20233d} has enabled real-time rendering through an explicit Gaussian ellipsoid representation and its differentiable splatting pipeline.
Despite these advances, all these methods produce degraded results when the number of images or their view coverage are limited.
To address these shortcomings, several works \cite{niemeyer2021regnerfregularizingneuralradiance, truong2023sparfneuralradiancefields, li2024dngaussianoptimizingsparseview3d,xiong2024sparsegsrealtime360degsparse} incorporate geometric constraints and depth regularization, leading to less artifacts and improving reconstruction fidelity under sparse-view conditions.
However, they remain insufficient for reconstructing explorable scenes, as they lack the ability to fill missing information in the training views.

\subsection{Evaluating novel view synthesis}
Previous novel view synthesis methods primarily focus on view interpolation, with LLFF \cite{mildenhall2019locallightfieldfusion}, Blender synthetic dataset \cite{mildenhall2020nerf}, TanksTemples \cite{TanksTemples} and MipNeRF360 \cite{barron2022mip} being widely used benchmarks.
These datasets place test viewpoints between training viewpoints and thus are not suitable for evaluating performance under large viewpoint changes.
To address this issue, Nerfbusters \cite{warburg2023nerfbustersremovingghostlyartifacts} introduces a new dataset with large viewpoint variations by capturing scenes from separate camera paths.
Nevertheless, most of the test views in Nerfbusters lie behind the training views, which are largely unconstrained and primarily require generating arbitrary contents rather than completing the under-reconstructed target scenes.

\subsection{Diffusion prior for 3D reconstruction}
Since optimization-based NVS methods cannot generate novel contents, recent works \cite{poole2022dreamfusiontextto3dusing2d, wang2022scorejacobianchaininglifting, wang2023prolificdreamerhighfidelitydiversetextto3d, liu2023zero1to3zeroshotimage3d, sargent2024zeronvszeroshot360degreeview} have explored methods for exploiting generative diffusion priors.
The vast majority of recent research \cite{gao2024cat3dcreate3dmultiview, shi2024mvdreammultiviewdiffusion3d, kong2024eschernet, huang2024epidiff, hu2024mvd, voleti2024sv3d, kwak2023vivid1to3, he2024cameractrl, wang2024motionctrl, liu2024reconx, liu20243dgsenhancer} train multi-view diffusion models or fine-tune video diffusion models on large multi-view datasets \cite{zhou2018re10k, ling2023dl3dv10k, reizenstein2021co3d} for generating 3D contents.
Among these, 3DGS-Enhancer \cite{liu20243dgsenhancer} fine-tunes a video diffusion model to enhance a sequence of 3D Gaussian splatting renderings to solve sparse-view reconstruction task.
On the one hand, some works \cite{shih2024extranerf, liu2024viewextrapolator} apply diffusion prior to address view extrapolation.
ExtraNeRF \cite{shih2024extranerf} introduces scene-level finetuned image inpainting diffusion and depth-completion model. 
ViewExtrapolator \cite{liu2024viewextrapolator} leverages SVD \cite{blattmann2023SVD} in a training-free manner.
While these methods are most closely related to our work, they assume known evaluation viewpoints and require running the diffusion model for every target view to render.
Moreover, their diffusion models lack multi-view consistency, leaving their ability of reconstructing explorable scenes unclear.

\subsection{Next-view selection for data acquisition}
A number of works studied methods to select the next viewpoint to acquire data when the number of input viewpoints are limited.
Early works focus on active learning or information-based strategies for efficient data acquisition \cite{shen2021stochastic, sünderhauf2022densityawarenerfensembles, pan2022activenerf, jiang2023fisherrf, inria2023vmv, xiao2024nerfdirector}. 
In particular, FisherRF \cite{jiang2023fisherrf} leverages the Fisher information matrix to quantify the information gain or uncertainty of a given viewpoint, serving as a strong baseline for next-best-view selection.
However, these methods primarily target single-object or forward-facing scenes, while our method demonstrates superior performance in complex scenes.
Recent works extend this line of works to autonomous exploration \cite{zhan2022activermap, jin2023neunbv, marza2024autonerf, liu2025riskaware}.
Instead of capturing additional inputs, ExploreGS samples virtual viewpoints from initially reconstructed scenes and exploits generative priors to synthesize images from that viewpoints.
\section{Method}
\vspace{-5pt}
\begin{figure*}[t]
    \centering
    \includegraphics[width=1\linewidth]{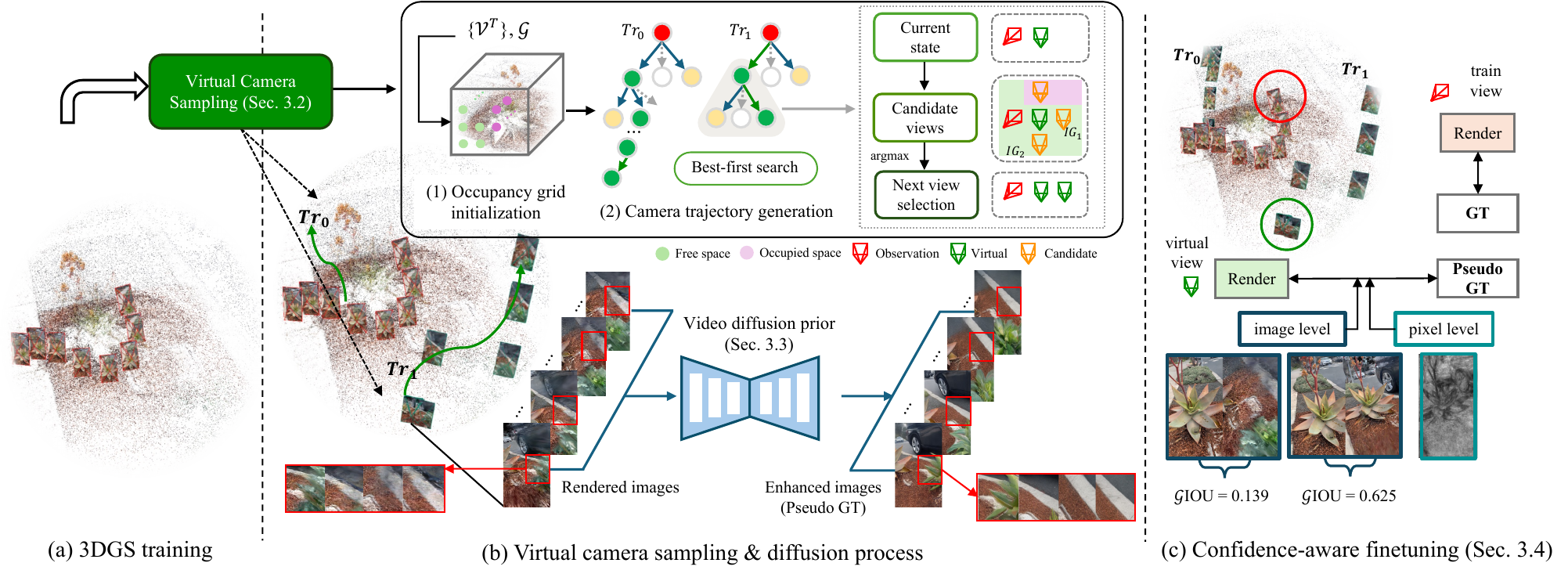}
    \vspace{-20pt}
    \caption{
    Overview of the proposed framework for scene exploration. 
    (a) The scene is initially optimized using 3DGS on the given training viewpoints. 
    (b) Based on the optimized 3DGS and training viewpoints, we generate virtual camera trajectories, and enhance the rendered views using our diffusion-based enhancement model. 
    (c) Finally, the scene is further optimized using both the original training viewpoints and the newly generated virtual viewpoints.
    }
   \label{fig:method}
   \vspace{-10pt}
\end{figure*}

\subsection{Overview}
The overall pipeline of ExploreGS is illustrated in \fref{fig:method}.
Our pipeline follows three stages: (1) scene initialization, (2) pseudo observation generation with virtual camera sampling and a video diffusion prior, and (3) fine-tuning 3DGS using both the training and the generated observations sets.

In the first stage, we begin by optimizing 3D Gaussians, denoted as $\mathcal{G}$, to represent the 3D scene using the given training viewpoints $\mathcal{V}^T=\{V^T_i\}^{N_T}_{i=1}$ where $N_T$ denotes the number of training viewpoints.
Here, each $V^T_i$ has known camera pose and ground truth image $I^T_i$.
Then, we determine the boundary of reconstructable scene based on the input observations and identify occupied regions for virtual viewpoints samplings.

In the second stage, we sample virtual camera trajectories based on information gain of each candiate virtual viewpoint, as shown in \sref{sec:virtual_view}.
$N_{Tr}$ camera trajectories that maximizes the information gain are sampled, and each trajectory $Tr_n$ consists of $L$ progressively shifting virtual viewpoints: $Tr_n = \{V^G_l\}_{l=1}^{L}$.
Along these virtual trajectories, we render images to obtain $I^G_i$ and generate pseudo-observations $I^P_i$ from $I^G_i$ by enhancing them using the generative prior of a video diffusion model.

Finally, we reconstruct complete and explorable 3D scenes by fine-tuning the initially optimized 3D Gaussians using both the training and virtual viewpoints, as described in \sref{sec:finetune}.
To mitigate the impact of inconsistencies between the generated and the original images during fine-tuning, we propose two techniques: (1) \textit{image-level confidence score}, and (2) \textit{pixel-level confidence map}.

\subsection{Scene initialization}
\label{sec:initialization}
\paragraph{Initial reconstruction.}
Our pipeline begins by optimizing the initial set of 3D Gaussians with the given training set within the 3DGS optimization framework \cite{kerbl20233d}.
Then, we estimate the target bounding box of the scene, which is a maximum boundary that can be reconstructed based on the given input images.
As previously discussed, reconstructing content beyond this bounding box is highly challenging, as it lacks grounding in the input observations and increasingly resembles unconstrained content generation, which is beyond the scope of this work.
This bounding box is computed by using the mesh extracted from the initial 3D Gaussians and the given cameras.

\paragraph{Rasterization-based occupancy estimation}
Since renderings from cameras inside occupied regions (\eg behind walls or inside solid objects such as an apple) offer redundant images, not relevant to the scene, they should be located in free space.
To this end, we introduce a simple rasterization-based algorithm to construct the occupancy grid $\mathcal{O} \in \mathbb{R}^{S \times S \times S}$.
We estimate the visibility of each 3D Gaussians using the Gaussian rasterizer by computing transmittance values from the training views.
Among these, we average the top$\text{-}$3 visibility scores to obtain the final visibility estimate.
A vertex is classified as occupied if the estimated visibility is below threshold $\tau$ (we use 0.5); otherwise, it is considered free space.

\begin{figure}[t]
    \centering
    \includegraphics[width=1\linewidth]{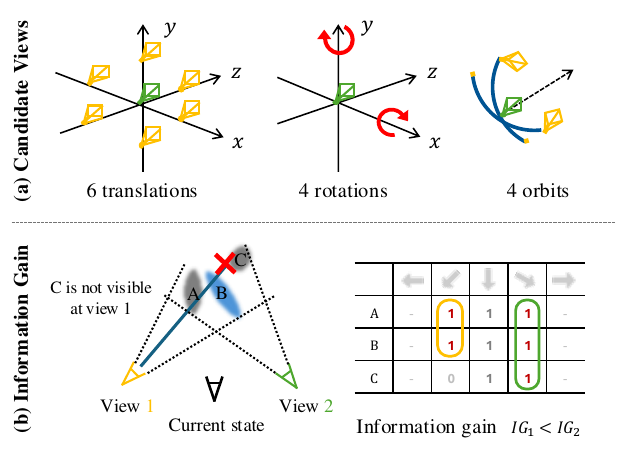}
    \vspace{-19pt}
    \caption{
    (a) Viewpoint candidates for virtual camera viewpoint generation. (b) Information gain of each viewpoint. Simplified 2D examples of both are presented for clarity.
    }
    \label{fig:camera}
    \vspace{-11pt}
\end{figure}
\subsection{Virtual view sampling}
\label{sec:virtual_view}
After initializing the target scene, our method utilizes video diffusion priors to supplement the missing information from the input views and removes unwanted artifacts and missing regions.
To effectively utilize diffusion priors, our sampling strategy selects virtual camera trajectories that can maximize the information gain of the scene.

As illustrated in \fref{fig:method}, our approach follows a best-first search strategy with depth-first characteristics to construct virtual trajectories.
We begin by selecting $N_{Tr}$ training viewpoints at spatially uniform intervals as the initial states for each trajectory and gradually select the next viewpoint.
At each step, the current viewpoint expands into $K$ candidate views. 
The one with the highest information gain becomes the next viewpoint in the trajectory, while the rest are stored for future consideration.
Here, the current state refers to the most recently added camera in a trajectory.

\paragraph{Candidate views.}
We define candidate views as the set of possible camera motion primitives, which determine the transitions from the current state. 
As shown in Fig.~\ref{fig:camera}, our action space consists of 6 translational moves, 4 pure rotations of the camera, and 4 rotations where the camera orbits around the look-at point, totaling 14 actions.
This choice ensure the virtual trajectory efficiently exploring the scene when compared to previous works just using going straight or turning around.

A candidate view is removed if it (1) lies outside the bounding box or in an occupied region, or (2) is close to the 3D Gaussians before fine-tuning, since rendering can be severely disrupted when the camera is placed near the occupied.
We utilize the target scene bounding box and the occupancy grid from \sref{sec:initialization} to achieve this.
These constraints prevent the trajectory from diverging away from the reconstructed scene or producing redundant pseudo-observations.

\paragraph{Information gain.}
We aim to observe 3D Gaussians densely from diverse viewpoints, as sparse or limited-angle observations often lead to inaccurate shape and color estimation.
To this end, we measure information gain based on view coverage.
We employ a 2D map $\mathcal{M} \in \mathbb{R}^{N_{\mathcal{G}} \times D}$, where $N_{\mathcal{G}}$ is the total number of Gaussians and $D$ is the number of discretized viewing directions around 3D Gaussians.
The map $\mathcal{M}$ stores binary values where each entry $\mathcal{M}[j,k] = 1$ indicates that the $j$-th Gaussian has been observed from the $k$-th viewing direction and initialized with training viewpoints $\mathcal{V}^T$.
At each step $t$, we update the map as follows:
\begin{equation}
    \mathcal{M}_t[j, k]=1, \quad \forall j \in J,
\end{equation}
where $J$ denotes the set of indices of 3D Gaussians visible from the candidate viewpoint and $k$ indicates the index assigned to the discretized view direction of the candidate viewpoint. 
This update is performed based on the previous map $\mathcal{M}_{t-1}$, which can be described as
\begin{equation}
    IG_t = \sum (\mathcal{M}_t - \mathcal{M}_{t-1}),
\end{equation}
where $IG_t$ is the number of Gaussians that have either been newly observed or newly viewed from different directions compared to $\mathcal{M}_{t-1}$.
3D Gaussians that are newly observed or seen from a different angle compared to $\mathcal{M}_{t-1}$, ensuring that only newly gained information is counted.
To avoid redundant overlaps among trajectories, $\mathcal{M}$ is shared across among all trajectories $Tr$.

\paragraph{Termination.} The search process terminates under two conditions. 
First, if no valid candidate views can be expanded from the current viewpoint while the number of cameras in the trajectory is still below the predefined length $L$, the current branch is pruned and the search resumes from the next-best candidate view in the priority queue.
Second, if the trajectory reaches the predefined length $L$, the search concludes. 
The resulting trajectory is treated as a set of pseudo-observations,
and then begins the searching process for the next trajectory $Tr_{i+1}$.
This process is repeated until $N_{Tr}$ trajectories are constructed.
We provide visualizations of example results in the supplementary materials.

\subsection{Diffusion prior}
\label{sec:diffusion}
Since renderings from virtual viewpoints often partially include well-reconstructed parts, it is desirable to leverage these visual cues.
To this end, we train a video diffusion model to enhance degraded rendered images $I^V$ to be realistic images $I^G$.
Our model is built on \cite{liu20243dgsenhancer} but introduces several key modifications.
First, unlike \cite{liu20243dgsenhancer}, which adopts a frame interpolation setup, we address extreme view extrapolation in scene exploration by performing next-frame prediction using only the observation closest to the virtual camera as the sole visual cue.
Second, to mitigate this ill-posed task, we condition the model on the scene text description and the ground truth image $I^T$ from the nearest training view to each pseudo-observation.
These are injected into the cross-attention layers of the diffusion model.
At last, we create a dataset specifically designed to model artifacts arising from extreme extrapolation.
The details about the dataset and the training process for the diffusion model are provided in the supplementary materials.

\subsection{Finetuning}
\label{sec:finetune}
Previous work  \cite{liu20243dgsenhancer} observed that applying an image-based loss (e.g., L1 loss) to a rendered image from a virtual camera and its pseudo ground truth often degrades well-reconstructed regions, as 3DGS is sensitive to minor inaccuracies from diffusion model.
It mitigates this issue by using two confidence levels, where higher confidence values indicate fewer conflicts between the contents of training and virtual viewpoints, thereby guiding 3D Gaussians to receive stronger gradients.
It adopts the image level confidence based on the distance between a virtual viewpoint and a training viewpoint and pixel level confidence based on the scale of 3D Gaussians.

Higher image-level confidence is applied when virtual viewpoints far from training viewpoints since they have less conflicts between them.
While effective, using distance alone often fails to account for pure camera rotations, as the distance remains unchanged even when the view frustum overlap between viewpoints decreases.
As our generated trajectories involve both translations and rotations, this results in suboptimal supervision.

To address this issue, we introduce a novel image-level confidence metric based on intersection-over-union (IoU) of Gaussians, denoted as $G\text{-IOU}$.
Our key insight is that a high view frustum overlap between a training viewpoint and a virtual viewpoint implies many co-visible 3D Gaussians, leading to more conflicts.
Using the Gaussian rasterizer, we efficiently extract the indices of visible Gaussians from each view, and define their intersection as the co-visible set:
\begin{equation} \mathcal{G}_{\text{covis}}(\mathcal{V}^{V}, \mathcal{V}^{\text{ref}}) = \mathcal{G}_{\text{vis}}(\mathcal{V}^V) \cap \mathcal{G}_{\text{vis}}(\mathcal{V}^{\text{ref}})
\end{equation},
where $\mathcal{G}_{vis}(\cdot)$ denotes the set of 3D Gaussians visible from a given viewpoint, and $\mathcal{V}_{ref}$ refers to the training viewpoint nearest to $\mathcal{V}^V$.
Based on this, $G\text{-IOU}$ is defined as
\begin{equation} 
    G\text{IOU}(\mathcal{V}^V, \mathcal{V}^{\text{ref}}) = \frac{|\mathcal{G}_{\text{covis}}(\mathcal{V}^{V}, \mathcal{V}^{\text{ref}})|}{|\mathcal{G}_{\text{vis}}(\mathcal{V}^{\text{ref}})|}
\end{equation},
and the image-level confidence is given by
\begin{equation} 
    U_{\text{img}} = 1 - G\text{IOU}(\mathcal{V}^V, \mathcal{V}^{\text{ref}})
\end{equation}.

For the pixel level confidence, rather than solely relying on the scale of 3D Gaussians which still produce artifacts depending on the viewpoint, we define it as the perceptual distance \cite{LPIPSmetric} between the rendered image $I^V$ and the output of the diffusion model $I^G$.
Each pixel in the confidence map reflects its contribution to the overall perceptual difference.
We upsample the pixel-aligned LPIPS map to the original image resolution and define pixel-level confidence as
\begin{equation}
    U_{\text{pixel}}(x,y) = \mathcal{U}\left(\text{LPIPS}\left(I^V, I^G\right)\right)(x,y)
\end{equation}

where $U_{\text{pixel}}(x, y)$ denotes the confidence value at pixel $(x, y)$ and $\mathcal{U}$ is an upsampling operation to the original image resolution.

Combining the above confidences, the loss for training views is defined as

\begin{equation}
    \mathcal{L}_{\mathcal{V}^T} = ||I^T-I^V||_1 + \mathcal{L}_{\text{D-SSIM}}(I^T, I^V)
\end{equation},
and the loss for virtual views is defined as
\begin{equation}
    \mathcal{L}_{\mathcal{V}^V} = U_{img} \cdot \left( ||I^G-I^V||_1 \odot U_{pixel} + \mathcal{L}_{\text{D-SSIM}}(I^G, I^V) \right)
\end{equation}.
The ${\odot}$ denotes the pixel-wise multiplication and $\mathcal{L}_{\text{D-SSIM}}$ denotes the dissimilarity variant of the SSIM loss.
This finetuning process is also illustrated in \fref{fig:method}~(c).

\begin{figure}[t]
    \centering
    \includegraphics[width=1\linewidth]{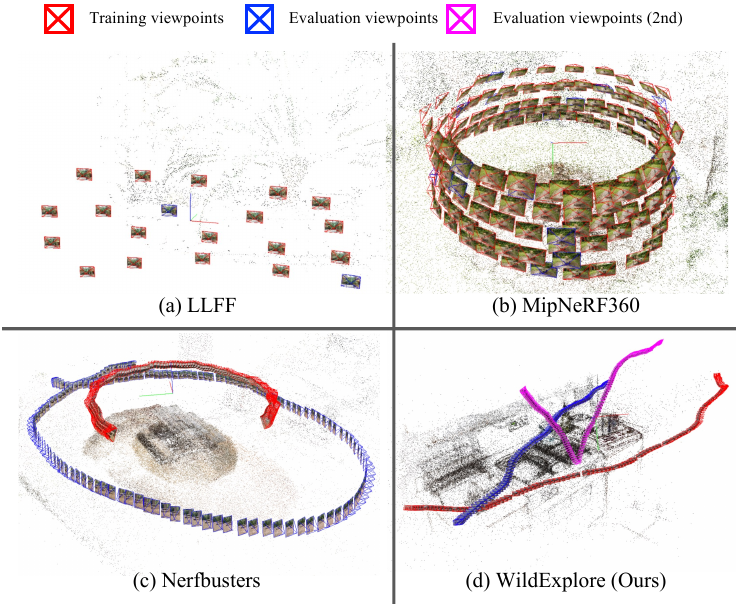}
    \vspace{-19pt}
    \caption{Viewpoint difference comparison with previous 3D reconstruction dataset. Training and evaluation viewpoints are visualized.}
    \label{fig:dataset}
    \vspace{-13pt}
\end{figure}
\section{Evaluation dataset for scene exploration}
\label{4Dataset}
Since existing datasets remain inadequate for evaluating scene exploration, as the test cameras are typically positioned within or close to the training trajectories, resulting in interpolation scenarios or only minor viewpoint variations.
Although the Nerfbusters dataset \cite{warburg2023nerfbustersremovingghostlyartifacts} introduces extreme novel viewpoints via two separate camera trajectories, its test views are placed far behind the training views. Consequently, the task aligns more with generative outpainting rather than complete scene reconstruction.
We introduce two datasets tailored for this task, with additional details provided in the supplementary materials.

\subsection{WildExplore} 
\label{WildExplore}
To address the lack of an appropriate benchmark for scene exploration, we introduce \textit{WildExplore}, a new dataset comprising four indoor and four outdoor scenes.
As illustrated in \fref{fig:dataset}, \textit{WildExplore} is designed with two key properties:
(1) it explicitly targets the exploration task by placing test viewpoints significantly distant from the training trajectories, ensuring large viewpoint transitions and broad scene coverage,
and (2) the evaluation set is captured to ensure sufficient co-visible areas between training and test views, minimizing the need for generating arbitrary contents, in contrast to Nerfbusters.

\subsection{Curated Nerfbusters}
\label{CuratedNerfbusters}
We curate the Nerfbusters dataset to better align with scene exploration objectives.
We selects some scenes where the test views are placed far behind the training views and swap the train/test split, while leaving the others unchanged.
This adjustment transforms the task into ``moving forward and closely examining the scene'', which better reflects realistic exploration scenarios.  
Our curated version includes seven scenes with swapped splits and two in their original form.
\section{Experiments}
\subsection{Experimental Setup}
\paragraph{Implementation details.}
We first train 3DGS using the given training viewpoints, following the original setup in \cite{kerbl20233d}.  
The number of virtual trajectories $N$ is set to 20 by default.
The occupancy grid resolution $S$ is set to 64.
At each step, we select the top 3 candidate views based on computed information gain.
We set $D$ as 32 for discretized view direction.
For the video diffusion model, we adopt DynamiCrafter \cite{dynamicrafter} as the backbone.
After generating the pseudo observation set, we fine-tune 3D Gaussians for 15K iterations, applying densification until 9K, with the same scheduling as 3DGS.
Further details will be discussed in the supplements.

\paragraph{Baselines.}
We evaluate our method in comparison to 3DGS-based methods, including 3DGS \cite{kerbl20233d}, 3DGS + Depth, DNGaussians \cite{li2024dngaussianoptimizingsparseview3d}, and ViewExtrapolator \cite{liu2024viewextrapolator}.
3DGS + Depth applies additional depth regularization using monocular depth estimation \cite{yang2024depthv2}, following the implementation from the official 3DGS repository \cite{kerbl20233d}.
DNGaussians \cite{li2024dngaussianoptimizingsparseview3d} incorporates depth priors during partial optimization of the 3D Gaussian parameters.
ViewExtrapolator leverages video diffusion but heuristically selects a single test camera for generating additional viewpoints due to the absence of a dedicated sampling strategy. 
In our experiments, we follow their protocol and choose the test camera farthest from the training set to encourage broader viewpoint coverage.

\begin{figure*}[t]
    \centering
    \includegraphics[width=1\linewidth]{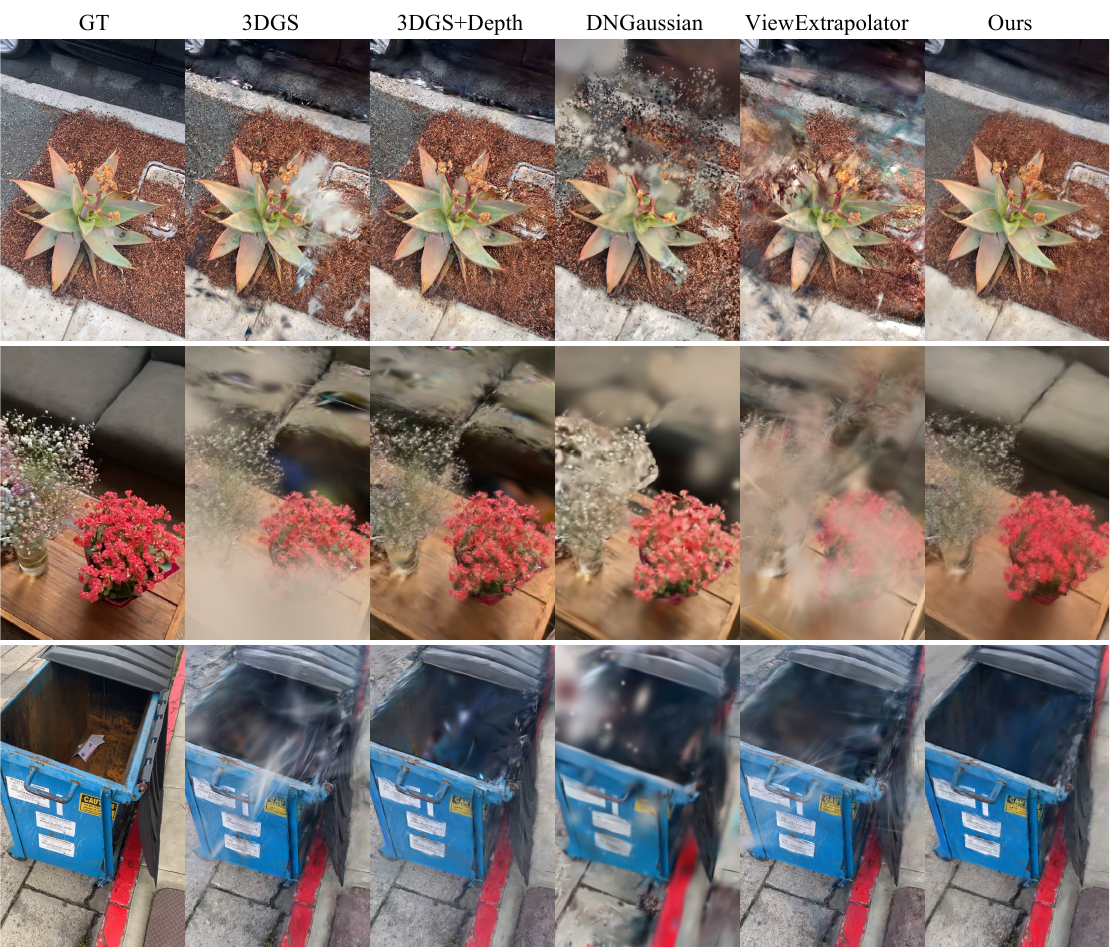}
    \vspace{-18pt}
    \caption{Qualitative comparison on the curated Nerfbusters dataset. }
    \label{fig:comparison_nerfbusters}
    \vspace{-13pt}
\end{figure*}

\begin{figure*}[]
    \centering
    \includegraphics[width=1\linewidth]{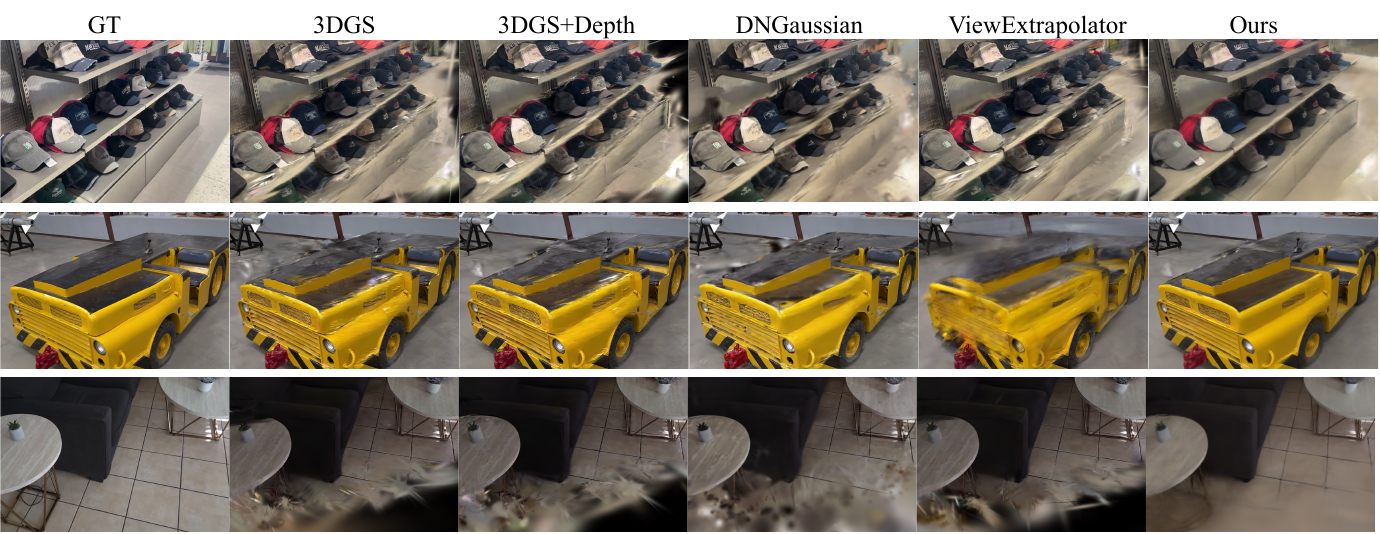}
    \vspace{-18pt}
    \caption{Qualitative comparison on the WildExplore dataset. }
    \label{fig:comparison_wildexplore}
    \vspace{-5pt}
\end{figure*}
\subsection{Results}
\paragraph{Quantitative comparison.}
\Tref{table:main} reports the quantitative results on the exploration task.
To assess reconstruction quality, we measure PSNR, SSIM, and LPIPS \cite{LPIPSmetric}.
The first row reports the result from 3DGS trained on both the training and test splits, serving as an upper bound.
Since our method effectively removes artifacts and fills missing regions, as observed in the qualitative analysis, our model outperforms across all metrics, particularly in PSNR and SSIM.

\paragraph{Qualitative comparison.}
\fref{fig:comparison_nerfbusters} and \fref{fig:comparison_wildexplore} show qualitative comparisons among our method and baseline methods.
3DGS \cite{kerbl20233d} suffers from artifacts, and its variants with depth regularization \changed{also meet the same problem, as they lack the capability to fill missing information}.
Although ViewExtrapolator \cite{liu2024viewextrapolator} utilizes a video diffusion prior and thereby shows competitive performance compared to 3DGS variants, it underperforms in challenging scenarios.
Its heuristic camera sampling struggles to maximize information gain and results in limited viewpoint coverage, leading to lower visual quality when rendering from distant viewpoints.
Its training-free diffusion prior also \changed{acts as a bottleneck}.
In contrast, our method fills missing regions and removes artifacts more effectively, producing images that align closely with the ground truth.
See the supplementary materials for additional results.

\begin{table}[t]
\setlength{\tabcolsep}{3pt}
\centering
\small
\resizebox{\columnwidth}{!}{
\begin{tabular}{c|ccc|ccc}
\toprule
\multirow{2}{*}{Method} & \multicolumn{3}{c|}{WildExplore} & \multicolumn{3}{c}{Curated Nerfbusters} \\ 
\cline{2-7} & PSNR$\uparrow$ & SSIM$\uparrow$ & LPIPS$\downarrow$ & PSNR$\uparrow$ & SSIM$\uparrow$ & LPIPS$\downarrow$ \\
\hline
3DGS Pseudo GT & 30.37 & 0.910 & 0.144 & 29.34	& 0.911	& 0.128 \\
\hline
3DGS & \cellcolor{yellow!40}15.81 & 0.487 & \cellcolor{yellow!40}0.414  & 14.51 & 0.417 & \cellcolor{yellow!40}0.454  \\
3DGS + Depth & 15.78 & 0.483 & \cellcolor{red!40}0.413 & \cellcolor{orange!40}15.00 &	0.435 & \cellcolor{orange!40}0.436 \\
DNGaussians & 15.65 & \cellcolor{yellow!40}0.504 & 0.450 & 14.01 & \cellcolor{yellow!40}0.442 & 0.525 \\ 
\hline
ViewExtrapolator &  \cellcolor{orange!40}16.02 &  \cellcolor{orange!40}0.510 & 0.442 & \cellcolor{yellow!40}14.98 & \cellcolor{orange!40}0.450 & 0.455 \\ 
\hline
Ours & \cellcolor{red!40}16.96 & \cellcolor{red!40}0.534 & \cellcolor{red!40}0.413 & \cellcolor{red!40}16.29 & \cellcolor{red!40}0.479 & \cellcolor{red!40}0.434 \\
\bottomrule
\end{tabular}}
\vspace{-5pt}
\caption{Quantitative comparison.}
\label{table:main}
\vspace{-5pt}
\end{table}

\begin{table}[t]
\renewcommand{\arraystretch}{1.2} 
\setlength{\tabcolsep}{10pt} 
\centering
\footnotesize
\resizebox{\columnwidth}{!}{
\begin{tabular}{c|ccc}
\toprule
\multicolumn{1}{c|}{Virtual View Sampling} & \multicolumn{3}{c}{Curated Nerfbusters} \\
\cmidrule(lr){1-4}
Info. Gain & PSNR$\uparrow$ & SSIM$\uparrow$ & LPIPS$\downarrow$ \\
\midrule
Grid view coverage \cite{marza2024autonerf} & 15.82 & 0.4675 & 0.4451 \\
FisherRF \cite{jiang2023fisherrf} & 15.77 & 0.4593 & 0.4394 \\
GS view coverage (Ours) & \textbf{16.29} & \textbf{0.4790} & \textbf{0.4336} \\
\bottomrule
\end{tabular}
}
\vspace{-5pt}
\caption{Comparisons on information gain design.}
\label{table:virtual_view_sampling}
\vspace{-5pt}
\end{table}

\begin{table}[t]
\renewcommand{\arraystretch}{1.2} 
\setlength{\tabcolsep}{7pt} 
\centering
\footnotesize
\resizebox{0.8 \columnwidth}{!}{
\begin{tabular}{c|ccc}
\toprule
\multirow{2}{*}{Method (K = 3)} & \multicolumn{3}{c}{\textit{Garbage}} \\
\cline{2-4} & PSNR$\uparrow$ & SSIM$\uparrow$ & LPIPS$\downarrow$ \\
\hline
3DGS & 14.42 & 0.480 & 0.526 \\
Ours with Bottom-K & 14.22 & 0.475 & 0.521 \\
Ours with Top-K & \textbf{15.45} & \textbf{0.526} & \textbf{0.508} \\ 		
\bottomrule
\end{tabular}
}
\vspace{-5pt}
\caption{Ablation study on information gain. TopK vs BottomK}
\label{table:vvsample_vs}
\end{table}
\begin{table}[t]
\setlength{\tabcolsep}{7pt}
\centering
\footnotesize
\resizebox{0.95\columnwidth}{!}{
\begin{tabular}{cc|ccc}
\toprule
\multicolumn{2}{c|}{Finetuning}& \multicolumn{3}{c}{Curated Nerfbusters} \\
\cmidrule{1-5} Image level & Pixel level & PSNR$\uparrow$ & SSIM$\uparrow$ & LPIPS$\downarrow$ \\
\cmidrule(lr){1-5}
Distance \cite{liu20243dgsenhancer} & - & 15.00 & 0.427 & 0.443 \\
- & Scale \cite{liu20243dgsenhancer} & 16.18 & 0.476 & 0.442 \\ 
Distance & Scale & 15.27 & 0.432 & \cellcolor{yellow!40}0.440 \\
\midrule
- & - & \cellcolor{red!40}16.30 & \cellcolor{yellow!40}0.476 & 0.447 \\
- & pixelLPIPS & \cellcolor{yellow!40}16.23 & \cellcolor{orange!40}0.478 & 0.441 \\
GS-IOU & -  & 16.10 & 0.473 & \cellcolor{orange!40}0.439 \\
GS-IOU & pixelLPIPS & \cellcolor{orange!40}16.29 & \cellcolor{red!40}0.479 & \cellcolor{red!40}0.434 \\
\bottomrule
\end{tabular}
}
\vspace{-5pt}
\caption{Ablations study on finetuning methods.}
\label{table:finetuning}
\vspace{-4pt}

\end{table}

\subsection{Ablation study}
\paragraph{Information gain.}
We validate the effectiveness of our information gain design by comparing it with alternative strategies, \changed{as shown in \Tref{table:virtual_view_sampling}.}
One alternative samples the 3D space uniformly and evaluate view coverage based on the occupied vertices of an occupancy grid, \changed{as proposed in \cite{marza2024autonerf}.}
Another follows typical active-learning strategies and select viewpoints with the highest uncertainty.
We adopt the method proposed in \cite{jiang2023fisherrf} for comparison.
As presented in the first and second rows, both show suboptimal results.
Grid-based approach often fails to maximize information gain, as it includes the gain from free space, resulting in redundant viewpoint selections. 
For uncertainty-based method, we observe that it is sensitive to 3D Gaussians with large scales, tending to chase them rather than addressing artifacts from viewpoint changes.

\paragraph{Search strategy.}
\changed{We ablate our best first search strategy to validate its effectiveness.}
As shown in ~\Tref{table:vvsample_vs}, the strategy selecting bottom K makes virtual cameras tend to stay near the trajectory of training viewpoints rather than deviating from it, leading to no performance gain.

\paragraph{Finetuning methods.}
We validate our finetuning approach by benchmarking it against the methods proposed by \cite{liu20243dgsenhancer} \changed{, as shown in \Tref{table:finetuning}}.
The model without the image-level confidence produces worse results in terms of LPIPS, as generated images whose viewpoints closely match the training set negatively impact already well-reconstructed regions.
Ignoring camera rotation further degrades performance, as observed in the second and fifth rows.
Although both our image-level confidence map and scale based one are effective, \changed{our method leads to slight better performance}.
\section{Conclusion}
This paper presents a framework that enables high-quality rendering of the novel view in arbitrary viewpoints.
Our method incorporates the video diffusion priors and the real-time rendering capability of 3DGS.
It introduces a sampling strategy to effectively sample trajectories of virtual cameras for maximizing information gain.

\textblock{Limitations and future works.}
While our method incorporates video diffusion priors to fill the missing information, we expect that employing more strong diffusion models \changed{which can faithfully generate high frequency details may} improve the overall performance.
In addition, extending the scene bounding box to cover a large scale scene would be an interesting avenue for the future work.
\clearpage

\section*{Acknowledgments} 
This work was supported by the Institute of Information \& Communications Technology Planning \& Evaluation (IITP) grant funded by the Korea government (MSIT) (No. RS-2022-II220113, No. RS-2024-00457882), Artificial Intelligence Graduate School Program, Yonsei University, under Grant RS-2020-11201361.
{
    \small
    \bibliographystyle{ieeenat_fullname}
    \bibliography{main}
}

\end{document}